\documentclass{article}
\usepackage{authblk}
\usepackage[utf8]{inputenc}
\usepackage[margin=1in]{geometry}
\usepackage{multicol}
\usepackage{multirow}
\usepackage{xcolor}
\usepackage{graphicx}
\usepackage{cite}
\usepackage{adjustbox}
\usepackage{float}
\usepackage{amsfonts}
\usepackage{url}
\usepackage{float}
\usepackage{bm}
\usepackage{braket}
\usepackage{amsmath}

\title{Learning Complex Word Embeddings in Classical and Quantum Spaces}

\author{Carys Harvey$^\ddag$\footnote{Work performed while at Quantinuum.}, Stephen Clark$^\dag$, Douglas Brown$^\dag$, and Konstantinos Meichanetzidis$^\dag$\\
$^\dag$Quantinuum\\
17 Beaumont Street, Oxford, UK\\
$^\ddag$Quantum Engineering Centre for Doctoral Training\\University of Bristol, UK\\
\hspace*{-0.4cm}\texttt{\small \{steve.clark,douglas.brown,kmei\}@quantinuum.com, carys.harvey@bristol.ac.uk}}

\date{\today}

\begin{document}

\maketitle

\begin{abstract}
We present a variety of methods for training complex-valued word embeddings, based on the classical Skip-gram model, with a straightforward adaptation simply replacing the real-valued vectors with arbitrary vectors of complex numbers. In a more ``physically-inspired'' approach, the vectors are produced by  parameterised quantum circuits (PQCs), which are unitary transformations resulting in normalised vectors which have a probabilistic interpretation. We develop a complex-valued version of the highly optimised C code version of Skip-gram, which allows us to easily produce complex embeddings trained on a 3.8B-word corpus for a vocabulary size of over 400k, for which we are then able to train a separate PQC for each word. We evaluate the complex embeddings on a set of standard similarity and relatedness datasets, for some models obtaining results competitive with the classical baseline. We find that, while training the PQCs directly tends to harm performance, the quantum word embeddings from the two-stage process perform as well as the classical Skip-gram embeddings with comparable numbers of parameters. This enables a highly scalable route to learning embeddings in complex spaces which scales with the size of the vocabulary rather than the size of the training corpus. In summary, we demonstrate how to produce a large set of high-quality word embeddings for use in complex-valued and quantum-inspired NLP models, and for exploring potential advantage in quantum NLP models.

\end{abstract}

\section{Introduction}

The neural revolution in NLP arguably began around a decade ago with the word embedding models of word2vec and GloVe \cite{mikolov-neurips-2013,pennington-etal-2014-glove}, followed by the successful application of recurrent neural networks (RNNs) to language tasks (especially LSTMs \cite{LSTM}) and finally the now-ubiquitous Transformer architecture \cite{attention}. Machine learning (ML) models based on tensor networks have also been proposed \cite{tensor-networks}, and researchers have begun to investigate whether quantum spaces can be utilised for modelling the rich probabilistic structures found in language \cite{sequence_processing_carys}. Over a similar period we have also seen a quantum computing revolution. While current quantum devices are still noisy \cite{preskill-nisq}, rapid improvements are now driving the field into the error-corrected, fault-tolerant regime \cite{moses2023race,daSilva2024} where algorithms with asymptotic speed-up over their classical counterparts can be successfully run \cite{nielsen2000}.

There have already been proposals for quantum RNNs, e.g. \cite{bauschQRNN,xuQRNN}, and quantum Transformers, e.g. \cite{khatriQTRANS,guoQTRANS}. In this report we focus on complex-valued word embedding models, for which the motivation is two-fold: first, because they provide a different geometry to their real-valued counterparts, which may provide advantages in learning lexical relations beyond semantic similarity, such as hierarchical relationships \cite{nickelPoincare}; and second, because normalised complex vectors are representations of quantum states \cite{aaronson-book}. In the latter case we prepare such representations using parameterised quantum circuits (PQCs), which are commonly used in the quantum machine learning (QML) literature. In this more ``physically-inspired'' approach, the vectors are produced by unitary transformations which result in normalised vectors and which come with a probabilistic interpretation.
ML training in large tensor spaces is in general hard, due to a phenomenon in QML known as ``barren plateaus'' \cite{barren_plateau}. Here we side-step this issue by training word representations in a reduced space in classical simulation. However, we note that there is still potential to utilise the full richness of tensor space during inference, by combining these atomic representations under some compositional structure \cite{qdiscocirc}. We also note the potential for advantage in sampling from structured models such as tensor networks \cite{huggins_2019}.

We present a variety of methods for training complex word embeddings, based on the classical Skip-gram model \cite{mikolov-neurips-2013}. For our first set of methods we simply replace the vectors of real values with vectors of complex numbers. At the heart of the Skip-gram (with negative sampling) method is the calculation of an inner product between the vector representation of a ``focal'' word and that of a context word, which is fed into a sigmoid to determine if the focal-context word pair actually occurred in some corpus. For the complex-valued case, the inner product is replaced with a (scaled) overlap measure which is the (scaled) absolute value of the inner product squared. One potential advantage of this measure is that, for normalised vectors, it lies in the interval $[0,1]$ and hence can be treated as a probability, and we investigate using this directly as the binary cross-entropy loss.

We provide two implementations based on arbitrary complex embeddings. The first is in PyTorch where each vector of complex numbers is represented by two arrays of real numbers: one for the real part of each complex number and one for the imaginary part.  Calculation of gradients is then handled automatically by PyTorch's autograd functionality (based on the real-number representations). The second implementation is an adaptation of the C code version of word2vec, where the gradients are calculated explicitly and used as part of a custom-made implementation of stochastic gradient descent. The advantage of the C code implementation is that it is highly efficient, exploiting the use of multi-threading on CPUs to easily enable the processing of billions of words of text with vocabularies into the hundreds of thousands.

In our second set of methods the complex vectors are produced by PQCs, which define a unitary transformation. Unitary transformations preserve L2-normalisation in the complex space and produce vectors which come with a probabilistic interpretation. Here we train and test the PQCs in noiseless classical simulation, using two techniques for training the PQCs. The first trains the PQCs directly, by using the word PQCs  to calculate the inner products needed during training. Here both the focal and the context words can be represented by PQCs, or just the focal word (with the context word an arbitrary complex embedding). The second technique takes the (normalised) complex word representations from the c program described above, and directly fits a PQC for each word to the corresponding representation. Since this is an arbitrary complex vector, there is no guarantee that the PQCs will be able to reproduce the corresponding state, but in fact we find that with an expressive ansatz we are able to fit the vectors perfectly. The advantage of this technique is that the second fitting stage only scales with the size of the vocabulary, and so the substantial increase in training time from introducing the PQCs is not a barrier to using large training sets.

We evaluate the complex embeddings on a set of standard similarity and relatedness datasets, and for some models obtain results which are competitive with the Skip-gram classical baseline. In particular, we find that, while training the PQCs directly tends to harm performance, the quantum word embeddings from the two-stage process, which are trained on a 3.8B-word corpus with a vocabulary of over 400k, perform as well as the classical real-valued embeddings with comparable numbers of parameters. This demonstrates that our method can produce a large set of high-quality word embeddings for use in complex-valued and quantum-inspired NLP models, and also for exploring potential advantage in quantum NLP models.

The rest of the report is organised as follows. Section~\ref{sec:models} defines a number of complex-valued word embedding models, based on the classical Skip-gram model. We explain how physically-inspired variants can be implemented using PQCs, and provide two schemes for word encoding, one which uses a single PQC for the whole vocabulary (basis encoding) and one which uses a separate PQC for each word (arbitrary encoding). Section~\ref{sec:expts} describes our experiments, which include directly training the PQCs using arbitrary encoding, and presents results on some standard similarity and relatedness datasets. Section~\ref{sec:related} gives some pointers to related work. Section~\ref{sec:conclusion} considers which models could in principle be run on quantum devices, and  provides a number of directions for future work.

\section{Complex Word Embedding Models}
\label{sec:models}

In this section we describe the various complex embedding models we have experimented with, based on the classical Skip-gram (with negative sampling) model.

\subsection{Skip-gram with Negative Sampling}
\label{sec:sg_ns}

The original Skip-gram model \cite{first-word2vec} defined a multi-class classification problem in which, given a focal-context word pair (taken from a fixed context window either side of the focal word), the task is to the predict the context word given the focal word. This task operationalises the so-called \emph{distributional hypothesis} \cite{harris,firth}, since words that occur in similar contexts will end up with similar representations. However, calculating the normalisation constant for the softmax is expensive, especially for large vocabularies, and so Mikolov et al. \cite{mikolov-neurips-2013} defined an alternative binary classification problem in which the task is to decide whether the focal-context word pair actually occurred in some training corpus. The negative examples are obtained by sampling from the vocabulary, and hence the method is known as Skip-gram with negative sampling (based on the more theoretically-motivated technique of noise contrastive estimation \cite{nce}).

The binary classification model is a sigmoid function which takes as input the inner product between the embeddings for the actual focal and context word, and the negative inner product for the negative examples, resulting in the following binary cross-entropy loss (for a single focal word) \cite{mikolov-neurips-2013}: 

\begin{equation}
    - \log \sigma (v_f \cdot v_c) - \sum_{i = 1}^k \log \sigma (- v_f \cdot v_{\tilde{c}_i})
    \label{eqn:sg_loss}
\end{equation}

\noindent where $\sigma$ is the sigmoid function, $(v_f, v_c)$ are the embeddings for the focal-context word pair, and $\{\tilde{c}_i \}_i$ is the set of embeddings for the negative context words. A typical value for the number of negative examples $k$ is 5. The negative examples are sampled according to a unigram distribution raised to the 3/4rd power, which was found to work better than the unigram or uniform distributions. For further details of the method, such as the subsampling of frequent words, we refer readers to the paper.

For the complex-valued versions we first simply replace the real values in the embeddings with complex numbers, and normalise the vectors whenever calculating inner products. A natural, real-valued overlap measure $F$ for vectors of complex numbers is the absolute value of the inner product squared:
\begin{equation}
F(v_f,v_c) = \vert \langle v_f \vert v_c\rangle \vert^2 \in [0,1],\;\; \text{for}\; v_f, v_c \in \mathbb{C}^n, ||v_f|| = ||v_c|| = 1,
\label{eqn:overlap}
\end{equation}
\noindent where $n$ is the dimension of the vector space, and $\langle v_f \vert v_c\rangle = \sum_{j=1}^n \overline{v_{f_j}}v_{c_j}$.\footnote{$F$ because this is the formula for the \emph{Fidelity} measure for (pure) quantum states. $\overline{v_{f_j}}$ is the complex conjugate of the $j$th component of $v_f$.} Note that the second sigmoid in (\ref{eqn:sg_loss}) requires negative values as input in order to correctly classify the negative examples, and so we introduce a scaling factor $D$ which effectively ``pulls over to the left'' some of the overlap values:
\begin{equation} 
F_{D}(v_f,v_c) = D \; (2  \vert \langle v_f \vert v_c\rangle \vert^2 - 1) \in [-D,D],\;\; \text{for}\; ||v_f|| = ||v_c|| = 1,
\label{eqn:sg_scaled_norm}
\end{equation}
\noindent resulting in the following loss function:
\begin{equation}
    - \log \sigma (F_D(v_f, v_c)) - \sum_{i = 1}^k \log \sigma (- F_D(v_f, v_{\tilde{c}_i})).
    \label{eqn:sg_loss_FS}
\end{equation}

\noindent  The intuition is that, if the overlap measure is $1$, then the scaled version is $D$, whereas if the overlap measure is $0$, then the scaled version is $-D$. We have found that a value of around $D = 3$ works well in practice (presumably because $\sigma(3)$ is close to 1 and $\sigma(-3)$ is close to $0$).

One potential advantage of using this normalised overlap measure is that it can be treated as a probability and used as a replacement for the sigmoid in the binary cross-entropy loss:
\begin{equation}
    - \log p(v_f,v_c) - \sum_{i = 1}^k \log(1 - p(v_f,v_{\tilde{c}_i})), \;\; \text{where} \;\; p(v_f,v_c) = \vert \langle v_f \vert v_c\rangle \vert^2.
    \label{eqn:ce_loss}
\end{equation}
In Section~\ref{sec:expts} we show that, for normalised arbitrary complex-valued vectors (i.e. not produced by a PQC), the ``direct'' cross-entropy loss in (\ref{eqn:ce_loss}) performs as well as the log-sigmoid loss in (\ref{eqn:sg_loss}).

\subsection{Parameterising Quantum Spaces}
\label{sec:pqcs}

Section~\ref{sec:sg_ns} described arbitrary complex embeddings in which the real values of the classical word2vec embeddings are simply replaced with complex numbers. In this section we provide a more ``physically-inspired'' parameterisation in which the complex vectors are produced by parameterised quantum circuits (PQCs). In a PQC, the parameters are not the complex values themselves, but rather the real values associated with gates in the quantum circuit (for example the angle associated with a single-qubit rotation gate). We call this model ``physically-inspired'' since PQCs can be implemented on quantum devices (subject to certain practical constraints such as the native gate set associated with the device, the fidelity of the gates, the depth of the circuit, and so on).
The original motivation for PQCs was that, since the parameters are real values, they can be optimised as part of a classical optimisation loop, leaving the task of any quantum device to be simply running the circuit (e.g. to calculate the value of the loss function), rather than performing the optimsation itself \cite{vqe,PQCs-benedetti}.

A particular PQC is associated with an \emph{ansatz} which consists of a choice of gates acting on a fixed number of qubits.\footnote{For a very short introduction to PQCs for NLP, see Ref.~\cite{xuQRNN}. For an authoritative and extensive introduction to quantum computing, see Ref.~\cite{nielsen2000}.} In general we would like this ansatz to be expressive, meaning that it is able to represent a large part of the underlying vector space. However, we would also like the ansatz to provide an appropriate bias so that a good solution to our problem can be found via gradient-based training. Given an ansatz, the number of qubits $n$ (which determines the dimension of the vector space $2^n$) and the number of layers (repetitions of the ansatz structure) are hyperparameters of the model, and these determine the number of trainable parameters. This is in contrast to the classical Skip-gram model where the number of parameters (for each word) is simply equal to the dimension of the vector space.

\begin{figure}[t!]
    \centering
    \includegraphics[width=0.5\linewidth]{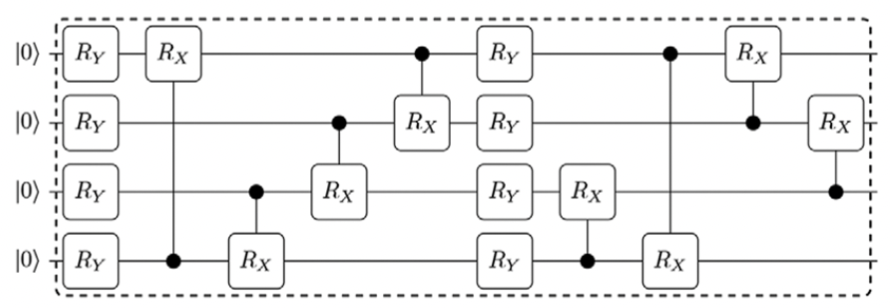}
    \caption{Ansatz 14 taken from \cite{expressive-pqc}.}
    \label{fig:a14}
\end{figure}

An example ansatz---``Ansatz 14'' from the circuit expressivity study in Ref. \cite{expressive-pqc}---is shown in Figure~\ref{fig:a14}. The input to the circuit is some easy-to-prepare state (typically the all-zero state, as is the case here), and the gates are either single-qubit $Y$-rotation gates, or two-qubit controlled $X$-rotation gates which create entanglement. Each of these gates has an angle associated with it, and the angles constitute the set of real-valued parameters to be optimised. Sim et al. demonstrate that a few layers of this ansatz are highly expressive, meaning that a large part of the vector space can be represented by the circuit. We have carried out extensive experimentation with Ansatz 14, and also Ansatz 5 (see Figure~\ref{fig:a5} in Section~\ref{sec:expts_pqc}), because they represent a reasonable trade-off in terms of expressivity and depth of circuit. The results in Section~\ref{sec:doug_expts} demonstrate that a few layers of these ansatze are sufficient for fitting PQCs to the arbitrary complex embeddings described in Section~\ref{sec:sg_ns}. 

The output of a PQC generally comes from taking measurements of the qubits in a particular basis. If all $n$ qubits are measured, then the output is a bit-string of length $n$ (from $2^n$ possibilities). The outcome of a measurement is probabilistic, and given by the \emph{Born rule}: the probability of a bit-string is the square of the magnitude of the \emph{amplitude} associated with that bit-string. The amplitudes are the complex numbers in the vector representation of the state (given a measurement basis). Here we perform all our experiments in classical simulation, so that expectation values of measurements can be calculated exactly, rather than as an empirical average over many samples or ``shots'', as would be required on a quantum device. In this report we also rely exclusively on inner products between states, in order to calculate the various $F$ measures from Section~\ref{sec:sg_ns}. The fidelity between two quantum states can be calculated using the circuit given in Figure~\ref{fig:fidelity_circuit} below, but here again we calculate the overlap analytically in classical simulation.\footnote{An alternative for calculating the fidelity on a quantum device is the swap test \cite{PhysRevLett.87.167902}, which trades off number of qubits for reduced circuit depth.}

\subsection{Encoding for the PQC}
\label{sec:pqc_encoding}

\begin{figure}[t]
\centering
\includegraphics[width=\textwidth]{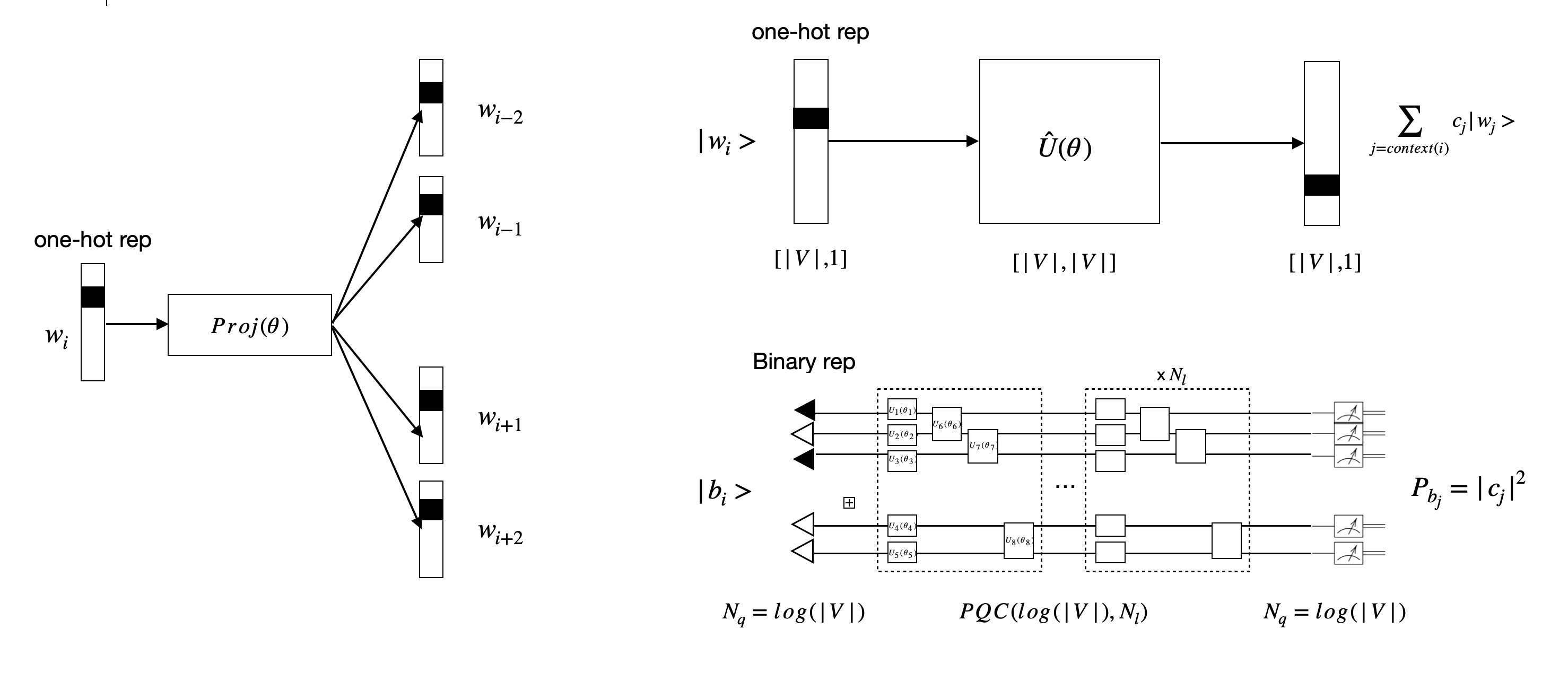}
\caption{Original Skip-gram (left), unitary implementation (right, top) and quantum circuit (right, bottom).}
\label{fig:encoding}
\end{figure}

Figure~\ref{fig:encoding} (left) shows the original Skip-gram model in which a distribution is defined over the whole vocabulary $V$, via a softmax applied to the focal word embedding (adapted from \cite{mikolov-neurips-2013}). A one-hot encoding of the focal word indexes into the word embedding matrix, which has dimension $[\vert V\vert, d]$ where $d$ is the arbitrary embedding dimension. So each row of that matrix is the embedding for a particular word. Even though we use Skip-gram with negative sampling in this report, and so never calculate the softmax, it is instructive to consider what a quantum version of this embedding matrix would look like, in order to motivate the possible encoding schemes for the PQC.

The diagram in the top-right shows a unitary version of the original Skip-gram model. The one-hot encoding picks out a column of the (complex) embedding matrix, but since that matrix represents a unitary transformation, it has to be square with dimensions $[\vert V\vert,\vert V\vert]$. The output for a particular focal word is a linear combination of basis vectors. The number of basis vectors is the same as the size of the vocabulary (or larger if the vocabulary size is not a power of 2), and the linear combination is normalised, and so the complex embedding for each word can be used to define a probability distribution over the possible context words (via the Born rule).

A PQC implementation of this scheme is shown in the figure (bottom right). There are two notable advantages to using a unitary transformation implemented with a PQC to define the embedding matrix. First, note that the number of qubits required to encode a one-hot input is only logarithmic in the size of the vocabulary (since $n$ qubits can encode $2^n$ bit-strings). Hence the $[|V|, |V|]$ unitary transformation is implemented as a PQC of width $\log(|V|)$ and arbitrary depth $N_l$. Second, the distribution over the context words comes ``for free'', rather than requiring the calculation of an expensive softmax, since the unitary transformation (by definition) maintains an L2-normalised state. There is a catch, however, which is that, in order to access the full distribution on an actual device, we would need to make $O(|V|)$ measurements; the underlying columns of the full $[|V|, |V|]$ matrix cannot simply be `read-off' like in the classical case.

\begin{figure}[t]
\centering
\includegraphics[width=0.8\textwidth]{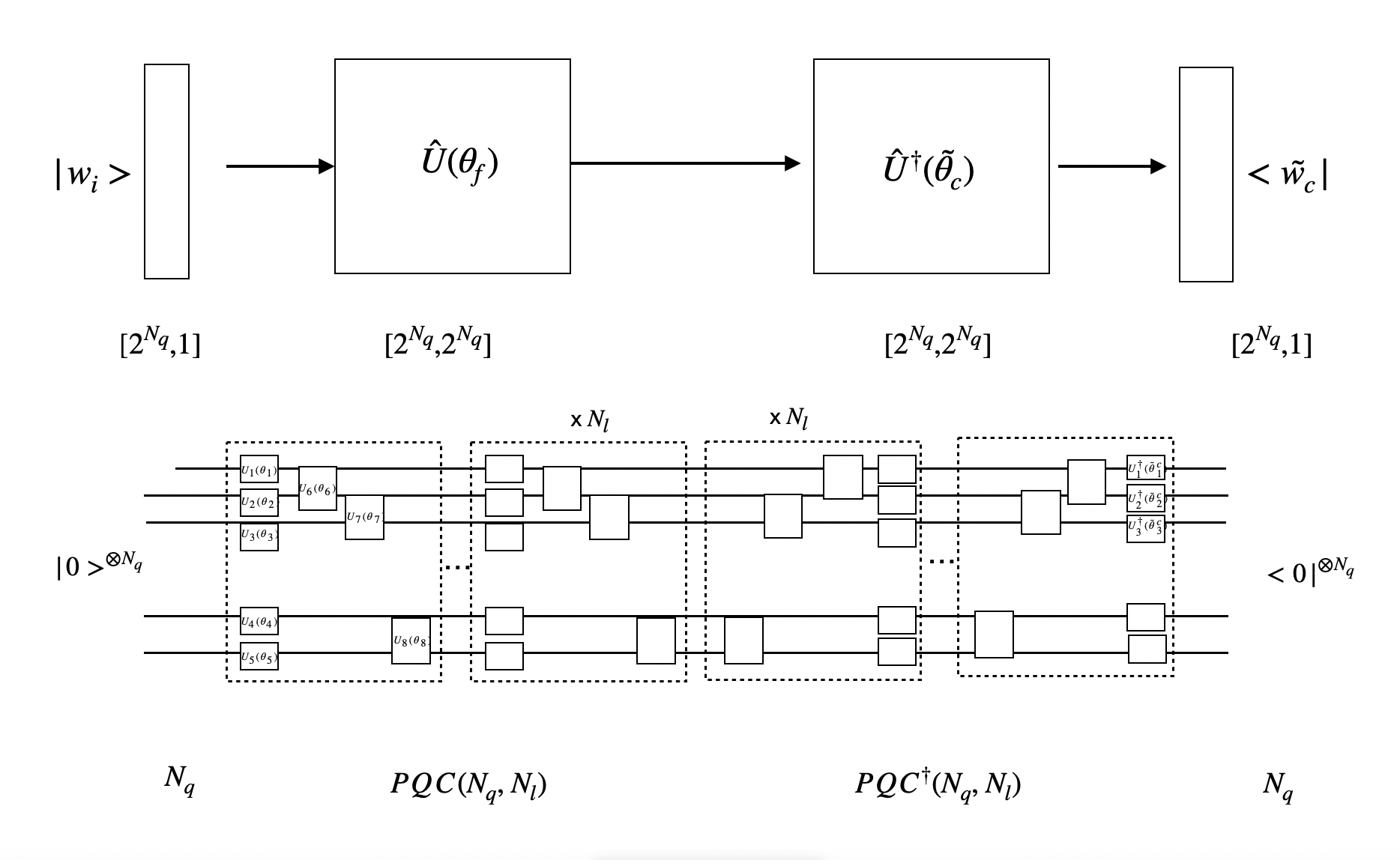}
\caption{Arbitrary encoding circuit for calculating the overlap between two quantum states.}
\label{fig:fidelity_circuit}
\end{figure}

The scheme described above is an example of \emph{basis encoding}, since each basis vector represents an element of the vocabulary. Note that a single unitary is used to encode the whole vocabulary. An alternative scheme, shown in Figure~\ref{fig:fidelity_circuit} and which we use in all the experiments below, is what we call \emph{arbitrary encoding}. Here a different unitary is used for each word; the circuit is the same, but each word has a different set of parameters. The advantage of this scheme is that the number of qubits is now an arbitrary-valued hyperparameter and no longer tied to the size of the vocabulary. This does mean that each embedding no longer defines a distribution over the vocabulary, so could not be used directly in the original Skip-gram model as in Figure~\ref{fig:encoding}, but for Skip-gram with negative sampling we just need to calculate overlaps between focal and context words.

\section{Experiments}
\label{sec:expts}

This section provides results for a variety of models on the WordsSim353 similarity and relatedness dataset \cite{ws353}. Section~\ref{sec:expts_complex} gives results for the arbitrary complex embeddings compared to the classical Skip-gram baseline, over a variety of embedding dimensions, as well as a comparison of the log-sigmoid loss (\ref{eqn:sg_loss_FS}) with using the overlap measure directly as the classification probability (\ref{eqn:ce_loss}). Section~\ref{sec:expts_pqc} gives results when using PQCs to prepare the complex word embeddings. Section~\ref{sec:scaling_up} provides results for the C code implementation of the complex Skip-gram model, and the corresponding PQCs which have been fitted to the output of this model. Section~\ref{sec:doug_expts} provides a comparison of a selection of ansatze from the circuit expressivity study in Ref. \cite{expressive-pqc}, showing how well various ansatze of varying depth are able to fit the arbitrary complex embeddings. Finally, Section~\ref{sec:more_eval} provides results on some additional similarity and relatedness datasets, demonstrating that the positive results for the complex models presented so far are not a result of overfitting to the WordSim353 dataset.

\subsection{Training, Datasets and Evaluation}
\label{sec:datasets}

We use two corpora for training the models. The first is the publicly available WikiText-103 corpus, containing just over 100M words from Wikipedia. The second, for the large-scale experiments in Section~\ref{sec:scaling_up}, is a larger Wikipedia dump which we have created ourselves, containing around 3.8B words. For the smaller corpus we apply a word frequency cutoff of 17, resulting in a vocabulary size of 101,529; and for the large-scale experiments we apply a cutoff of 100, resulting in a vocabulary size of 426,507.

The number of negative examples $k$ is set at 5 for all training runs, with a window size of 5 (i.e. 5 words either side of the focal word). For the PyTorch implementations, Adam is used for optimization, and the learning rate is 0.001 (unless stated otherwise). The scaling factor $D$ in (\ref{eqn:sg_scaled_norm}) is set at 3.5 for all experiments using loss (\ref{eqn:sg_loss_FS}).

The evaluation datasets contain word pairs together with a human-labelled similarity or relatedness score. The Spearman correlation is a measure of how well the ranking of the pairs from these scores correlates with the ranking from the system-assigned scores. For the classical baseline, the system-assigned score for a pair of words is the normalised inner product between the corresponding vectors (i.e. the cosine measure). For the complex models, the score is the (normalised) Fidelity measure from Section~\ref{sec:sg_ns}. 

For the WordSim353 dataset, the coverage for the larger vocabulary is 100\%. For the smaller vocabulary, there is one pair in WordSim353 with a word missing (which we ignore in the evaluation).

\subsection{Complex-valued Embeddings}
\label{sec:expts_complex}

\begin{table}[t!]
 \begin{center}
     \begin{tabular}{l|r|r}
     \hline
      Model & Embedding Dim & Correlation \\
      \hline\hline
      Real-valued Skip-gram  & 100 & $65.1_{\,0.22}$\\
      Complex Skip-gram w/sigmoid    & 64  & $64.4_{\,0.16}$  \\
      Complex Skip-gram w/direct prob    & 64  & $64.6_{\,0.33}$  \\
      \hline
\end{tabular}
\caption{Spearman correlation on WordSim353 for classical and complex Skip-gram, PyTorch implementation trained on WikiText-103. Averages  over 3 runs; SD in subscript.}
\label{tab:complex_sg}
\end{center}
\end{table}

\begin{table}[t!]
 \begin{center}
     \begin{tabular}{l|r|r}
     \hline
      Model & Embedding Dim & Correlation \\
      \hline\hline
      Classical Skip-gram  & 200 & $66.7_{\,0.50}$\\
                          & 100 & $65.1_{\,0.22}$\\
                          & 64 & $63.0_{\,0.50}$\\
                          & 32 & $58.6_{\,0.38}$\\   
                          & 16 & $52.4_{\,0.58}$\\   
                          \hline
      Complex Skip-gram w/direct prob    & 128  & $63.9_{\,0.24}$  \\
                                      & 64 & $64.6_{\,0.33}$\\
                                      & 32 & $63.9_{\,0.71}$\\
                                      & 16 & $60.0_{\,0.46}$\\   
                                      & 8 & $52.0_{\,0.86}$\\
      \hline
\end{tabular}
\caption{Spearman correlation on WordSim353 for different embedding sizes for classical and complex Skip-gram. Averages over 3 runs; SD in subscript. 10 epochs for training.}
\label{tab:emb_sizes}
\end{center}
\end{table}

For the classical baseline we use the Andras7 PyTorch implementation.\footnote{\url{https://github.com/Andras7/word2vec-pytorch}} This also forms the basis for our PyTorch implementation of the complex-valued version, which simply replaces each real-valued word embedding with two arrays of real numbers: one for the real part of the complex vector and one for the imaginary part. The overlap can then be calculated explicitly using these real representations without  having to use any complex types (although PyTorch does provide support for these), and gradients with respect to the real representations can be calculated automatically using PyTorch's autograd functionality.

Table~\ref{tab:complex_sg} gives results on the WordSim353 dataset for two complex-valued Skip-gram models, compared with the real-valued baseline, with the models trained on the smaller WikiText-103 corpus for 10 epochs. The correlation numbers are averages over 3 training runs (with different random seeds). The complex model with the sigmoid uses the loss in (\ref{eqn:sg_loss_FS}) and the model with the direct probability uses the loss in (\ref{eqn:ce_loss}). The complex models have 128 parameters per word (64 for the real part of each complex vector and 64 for the imaginary part), which corresponds to 6 qubits for the PQC models below. The training took around 10 minutes per epoch for the real-valued baseline, and around 15 minutes per epoch for the complex version (on a single A30 GPU with 24GB RAM). The summary is that the complex versions are competitive with the baseline, with roughly comparable numbers of parameters, and the model which uses the Fidelity directly as a classification probability performs as well as the version which feeds the scaled Fidelity into a sigmoid for classification.

Table~\ref{tab:emb_sizes} shows how the results vary for different embedding sizes, for the real-valued baseline and the complex model which uses the Fidelity measure as the classification probability. The drop-off in accuracy follows a similar trend for the two models for comparable numbers of parameters, although the increase in accuracy for the real-valued model when increasing the embedding dimension (from 100 to 200) is not mirrored in the complex case (when increasing from 64 to 128).

\subsection{PQC-based Embeddings}
\label{sec:expts_pqc}

\begin{table}[t!]
 \begin{center}
     \begin{tabular}{l|r|r}
     \hline
      Model & Embedding Dim & Correlation \\
      \hline\hline
      PQC (focal) Skip-gram w/sigmoid    & 64  & $63.5_{\,0.34}$  \\
      \hline
        PQC (focal) Skip-gram w/sigmoid   & 128  & $61.4_{\,0.26}$  \\
                                    & 32  & $62.9_{\,1.29}$  \\
                                    & 16  & $62.4_{\,0.43}$  \\
                                    & 8  & $53.3_{\,1.28}$  \\
      \hline
\end{tabular}
\caption{Spearman correlation on WordSim353 for complex Skip-gram with the focal word as a PQC, with different embedding dimensions. Averages over 3 runs; SD in subscript. 10 epochs for training.}
\label{tab:pqc_focal}
\end{center}
\end{table}

\begin{figure}[t!]
    \centering
    \includegraphics[width=0.5\linewidth]{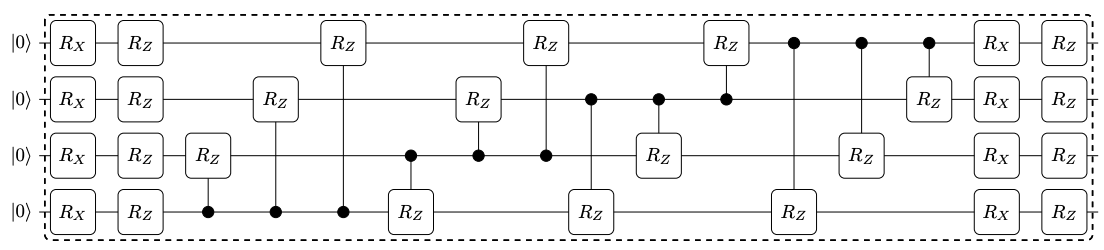}
    \caption{Ansatz 5 from \cite{expressive-pqc}.}
    \label{fig:a5}
\end{figure}

Table~\ref{tab:pqc_focal} gives results on WordSim353 for the case when the focal word embedding is produced by a PQC, and the context word embedding is an arbitrary complex vector. The number of qubits needed for an embedding dimension of $N$ is $\log_2(N)$. The circuit we used is 3 layers of Ansatz 5 (A5) from Ref. \cite{expressive-pqc}, which is shown in Figure~\ref{fig:a5} (with 4 qubits, i.e. an embedding dimension of 16).\footnote{Figs.~\ref{fig:a14} and \ref{fig:a5} show the all-zero state as input to the circuit, whereas for all the experiments in this report we first applied a Hadamard gate to each qubit so all basis states are in a uniform superposition, but essentially any easy-to-prepare state is suitable as input.} A5 uses a different set of rotation gates to A14 and has more parameters per layer. In particular, for A14 with a vector space of dimension $N$ and $l$ layers, the number of parameters is $4 \log_2(N) l$. For A5 it is $(\log_2^2(N) + 3 \log_2(N))l$. We found that the results for A5 for this experiment were marginally better than for A14.

Here we take inner products between the embedding produced by the PQC for the focal word and the normalised arbitrary complex embedding for the context word, and feed the scaled overlap into the sigmoid (\ref{eqn:sg_loss_FS}). Introducing a PQC substantially increases the training time, with each epoch taking roughly 1.5 hours (for 6 qubits). We also found that a larger learning rate (0.01) worked better for the PQC-based models. The results are comparable to those in Table~\ref{tab:emb_sizes} for the complex Skip-gram model with no PQCs, across all embedding dimensions.

Table~\ref{tab:pqc_focal_context} gives results for the model which uses a PQC for both the focal and the context word, for 3 layers of both A5 and A14. The PQC embeddings are already normalised, and the overlap measure is fed into the sigmoid as above. Introducing PQCs for the context words increases the training time further, since the time now scales with the number of negative context examples. We also found that more epochs are needed for training (20). Each epoch takes roughly 12 hours (for 3 layers of A5), and we find that introducing the additional PQC harms accuracy overall, with A14 performing better in this case than A5.

\begin{table}[t!]
 \begin{center}
     \begin{tabular}{l|r|l|r}
     \hline
      Model & Embedding Dim & Anzatze & Correlation \\
      \hline\hline
      PQC (focal, context) Skip-gram w/sigmoid    & 64  & A5 & $53.4_{\,0.10}$  \\
      PQC (focal, context) Skip-gram w/sigmoid    & 64  & A14 & $57.8_{\,0.41}$  \\
      \hline
\end{tabular}
\caption{Spearman correlation on WordSim353 for complex Skip-gram  with focal and context word as a PQC. Averages over 2 runs for A5, 3 for A14; SD in subscript. 20 epochs for training. }
\label{tab:pqc_focal_context}
\end{center}
\end{table}

\subsection{Scaling Up and Mapping to PQCs}
\label{sec:scaling_up}

We have described a Skip-gram method for obtaining a set of high-quality word embeddings produced by PQCs, using arbitrary complex vectors for the context words (see Table~\ref{tab:pqc_focal}). However, even when we only use PQCs for the focal word, the training times are becoming expensive for a modest corpus size of 100M words. The bottleneck occurs when a PQC is needed to calculate inner products for each word token in the training corpus. Here we describe a method which only scales with the size of the vocabulary rather than the training corpus, and which can easily scale to billions of words of training data and vocabulary sizes in the hundreds of thousands. The large Wikipedia corpus and vocabulary used in the experiments were described in Section~\ref{sec:datasets}. 

The motivation for scaling up is provided by the original c implementation of the word2vec models, which exploits multi-threading on CPUs to provide a highly efficient implementation of Skip-gram with negative sampling.\footnote{\url{https://github.com/tmikolov/word2vec/blob/master/word2vec.c}} It takes less than 1 hour per epoch for training on the large corpus (on a standard Linux server with 10 threads). Gradients are calculated explicitly in the code and fed into a custom-made implementation of mini-batch stochastic gradient descent. In order to create a complex-valued version, we represent each complex word embedding as two arrays of real numbers: one for the real part and one for the imaginary part, as for the PyTorch implementation described earlier. However, rather than rely on PyTorch's autograd functionality to calculate gradients, we now have to calculate these explicitly in the C code. One difference with the complex model from Section~\ref{sec:sg_ns} is that we have not normalised the complex vectors before calculating the overlap (since this simplifies the gradient calculations)\footnote{We have tried a normalised version but did not see any improvement in the results.}, and so a slightly different scaling is required before feeding into the sigmoid for the loss function:
\begin{equation} 
F_D(v_f,v_c) = 2  \vert \langle v_f \vert v_c\rangle \vert^2 - D \in [-D,\infty).
\end{equation}
\noindent The intuition is similar to before: if the overlap measure is $D$ then the scaled version is still $D$, whereas if the overlap measure is 0 then the scaled version is $-D$. In the experiments we used a value of $D = 3.5$ (as before).

\begin{table}[t!]
 \begin{center}
     \begin{tabular}{l|r|r}
     \hline
      Model & Embedding Dim & Correlation \\
      \hline\hline
      Classical Skip-gram  & 100 & $72.2_{\,0.25}$\\
      Complex Skip-gram w/sigmoid    & 64  & $68.8_{\,0.45}$  \\
      \hline
      Complex Skip-gram (one run)    & 64  & $69.3$  \\
      Corresponding fitted PQCs                   & 64  & $69.3$  \\
      \hline
\end{tabular}
\caption{Spearman correlation on WordSim353 for classical and complex Skip-gram, c implementation trained on larger dataset (top). Averages  over 3 runs; SD in subscript. Results for the fitted PQCs (bottom).}
\label{tab:c_impl}
\end{center}
\end{table}

Table~\ref{tab:c_impl} shows results on WordSim353 for the real-valued baseline and the complex equivalent, for 2 epochs of training (which is still only taking around 1 hour per epoch for the complex version). Even though the complex version is a few percentage points below the baseline, the results are a significant improvement over the earlier results on the smaller corpus. 

Now we have a method for producing a set of high-quality complex word embeddings, but these could not be used, for example, in a QNLP experiment utilising a quantum device. The reason is that an exponential circuit depth is required to prepare an arbitrary quantum state (unless ancillary qubits are used, but in the
worse case the number of ancillas also scales exponentially) \cite{PhysRevLett.129.230504}. Hence in a second stage we fit a PQC for each word to the corresponding (normalised) complex word embedding. The loss is simply the overlap measure from (\ref{eqn:overlap}), where $v_f$ is the complex vector corresponding to the state prepared by the PQC (for the word in question), and $v_c$ is the corresponding normalised complex vector from the c implementation in stage 1. The fitting code is implemented in PyTorch using the Adam optimizer (with a learning rate of 0.01). The training data was split into 3 chunks, with each chunk fitting on a single A30 GPU with 24GB RAM, and the training time was on the order of a few hours (for the larger chunks and for 5,000 iterations).

It is an interesting question whether a particular ansatz and number of layers can provide a good fit to these arbitrary vectors, given the hardness of preparing an arbitrary quantum state. In fact we find that, with e.g. 3 layers of A5, and 5,000 iterations of training, we can fit the vectors perfectly, with the loss going to effectively zero. The bottom half of Table~\ref{tab:c_impl} shows the best result on WordSim353 from the 3 runs used to get the complex Skip-gram results in the top half, and the result for the PQCs which have been fitted to the embeddings from that run. The PQC for each word was 3 layers of A5 and, since the loss effectively went to zero, the performance of the PQCs on WordSim353 was identical to the arbitrary complex embeddings on which the PQCs were trained. The significance of this result is that we now have PQCs for a large vocabulary of over 400k words which can be used to prepare a set of high-quality quantum word embeddings.

\subsubsection{Comparing Anzatze and Number of Layers}
\label{sec:doug_expts}

\begin{figure}
    \centering
    \includegraphics{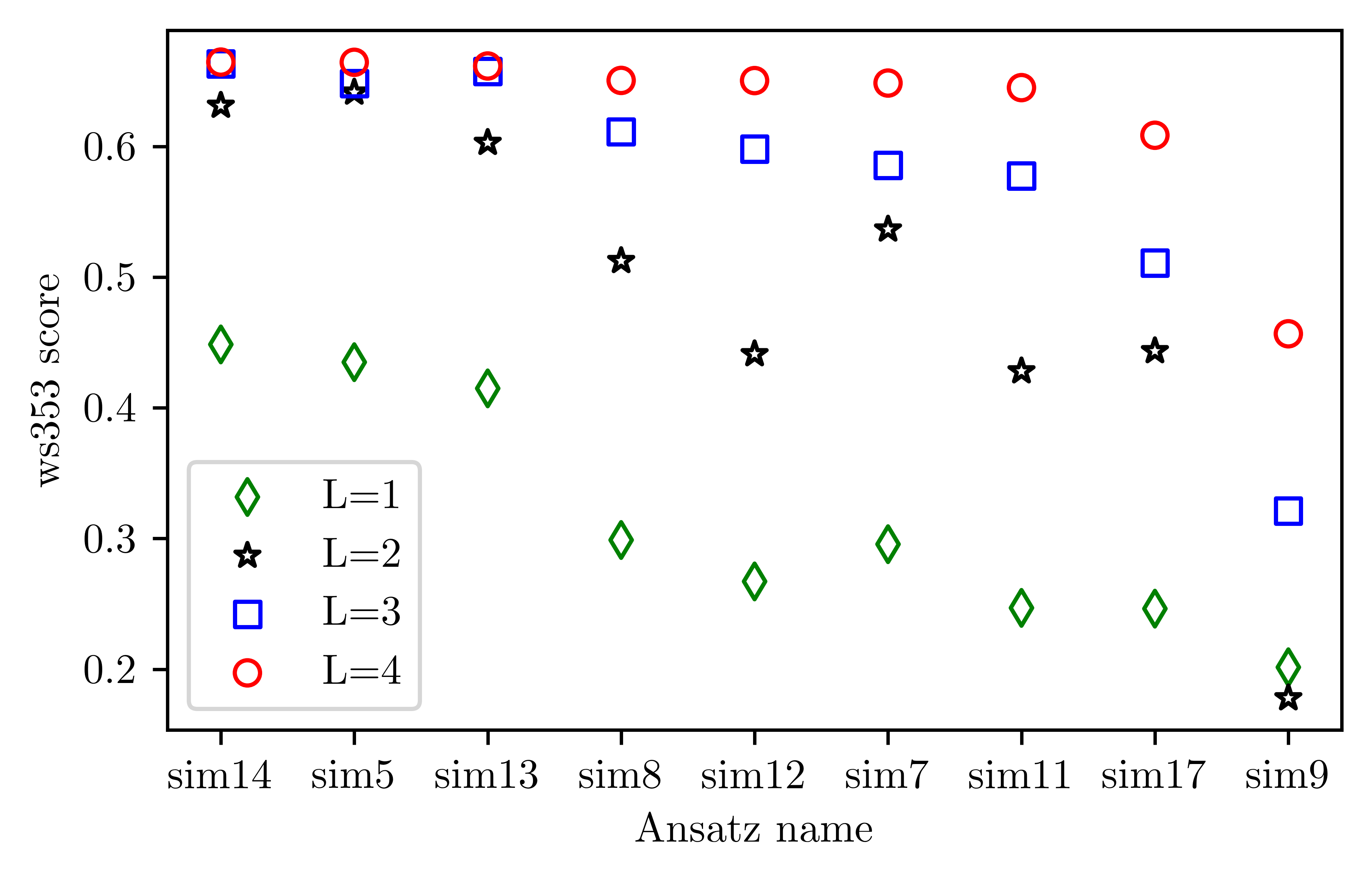}
    \caption{WordSim353 scores for a variety of ansatze from Sim et al. with 6 qubits and L = no. of layers.}
    \label{fig:ansatz_ws353}
\end{figure}

Now that we have an efficient method for training PQCs, this setting provides an excellent test bed in which to compare various ansatze and number of layers, in order to investigate what is required for fitting the arbitrary complex vectors from stage 1. Figure~\ref{fig:ansatz_ws353} shows the WordSim353 results for a variety of ansatze and number of layers (from 1 to 4), with the circuits taken from Ref. \cite{expressive-pqc} (see Appendix~\ref{sec:ansatze} for the ansatze). Sim et al. propose various descriptors which provide measures of expressibility and entangling capability of circuits, which can be statistically estimated from classical simulations of PQCs. We can think of our study as mirroring that of Sim et al., but with the task of measuring expressibility replaced with our particular use case of fitting PQCs to arbitrary complex word embeddings. 

For this experiment, the arbitrary complex word embeddings used for fitting the PQCs were obtained by running the c implementation described above on the smaller WikiText-103 corpus, for 5 epochs, with a frequency cutoff of 17 (giving a vocabulary size of 101k), and an embedding dimension of 64 (corresponding to 6 qubits). The PyTorch fitting code was run for 10,000 iterations, with a learning rate of 0.01. The plot in Figure~\ref{fig:ansatz_ws353} shows some strong  similarities with the corresponding plot in Figure 3 of Ref~\cite{expressive-pqc}, with A5 and A14 performing well overall, and with 3 to 4 layers needed in general for good performance.

\subsection{Results on Additional Similarity Datasets}
\label{sec:more_eval}

Our final set of experiments uses three more similarity and relatedness datasets, to ensure that the positive results we have presented so far are not due to overfitting to the WordSim353 dataset. The datasets are: MEN\footnote{\url{https://staff.fnwi.uva.nl/e.bruni/MEN}}, RG-65\footnote{\url{https://aclweb.org/aclwiki/RG-65_Test_Collection_(State_of_the_art)}}, and SCWS \cite{huang-etal-2012-improving}; and each contains some number of word pairs, from various sources, with each pair assigned a human semantic similarity or relatedness score. 

Table~\ref{tab:more_data_1} shows results for the various PyTorch models trained on the WikiText-103 corpus, using the single set of weights for each model which performed best on the WordSim353 dataset. The embedding dimension is 64 for the complex models (6 qubits for the PQCs), and 100 for the classical real-valued baseline. The 1-way PQC model uses 3 layers of A5 for the focal word and an arbitrary complex embedding for the context word; 2-way uses 3 layers of A14 for both focal and context words. The word pair coverage with the 101k vocabulary is 99.7\% for WordSim353, 98.6\% for MEN, 93.8\% for RG-65, and 95.3\% for SCWS (with the missing pairs being ignored in the evaluation). Overall the results are repeated from earlier, with the complex models being competitive with, and in some cases outperforming, the real-valued baseline (including the 1-way PQC model which uses a PQC for the focal word). The only model which is consistently below the baseline is the 2-way PQC model.

\begin{table}[t!]
 \begin{center}
     \begin{tabular}{l|l|r}
     \hline
      Model & Dataset (\# pairs) & Correlation \\
      \hline\hline
      Classical Skip-gram  & WordSim353 (353) & $65.4$\\
      Complex Skip-gram w/sigmoid  &   & $64.6$  \\
      Complex Skip-gram w/direct prob  &   & $65.0$  \\
      Complex Skip-gram 1-way PQC  &   & $64.0$  \\
      Complex Skip-gram 2-way PQC  &   & $58.3$  \\
      \hline
      Classical Skip-gram  & MEN (3,000) & $62.7$\\
      Complex Skip-gram w/sigmoid  &   & $65.0$  \\
      Complex Skip-gram w/direct prob  &    & $64.0$  \\
      Complex Skip-gram 1-way PQC  &   & $62.0$  \\
      Complex Skip-gram 2-way PQC  &   & $57.9$  \\
      \hline
      Classical Skip-gram  & RG-65 (65) & $64.6$\\
      Complex Skip-gram w/sigmoid  &   & $70.5$  \\
      Complex Skip-gram w/direct prob  &    & $69.7$  \\
      Complex Skip-gram 1-way PQC  &   & $69.4$  \\
      Complex Skip-gram 2-way PQC  &   & $56.6$  \\
      \hline
      Classical Skip-gram  & SCWS  (2,003) & $63.5$\\
      Complex Skip-gram w/sigmoid  &   & $62.6$  \\
      Complex Skip-gram w/direct prob  &   & $61.8$  \\
      Complex Skip-gram 1-way PQC  &   & $61.4$  \\
      Complex Skip-gram 2-way PQC  &   & $59.8$  \\
      \hline
\end{tabular}
\caption{Spearman correlation on various datasets for the PyTorch models with an embedding dimension of 64 for complex, 100 for real-valued; 3 layers of A5 for 1-way PQC, A14 for 2-way.}
\label{tab:more_data_1}
\end{center}
\end{table}

Table~\ref{tab:more_data_2} shows the results for the large-scale models trained on the 3.8B-word corpus. The word pair coverage with the 426k vocabulary is 100.0\% for WordSim353, 99.7\% for MEN, 100.0\% for RG-65, and 98.8\% for SCWS. The first two rows for each dataset are from the c implementations, and the fitted PQC in the third row is from the PyTorch fitting code, with 3 layers of A5. Again WordSim353 is used as a validation set to choose the best-performing model to run on the additional datasets. The complex Skip-gram model is again competitive with the real-valued baseline, and the fitted PQC matches the arbitrary complex vectors from the second row for each dataset exactly, except for the SCWS dataset where it is 0.2 percentage points below.

\begin{table}[t!]
 \begin{center}
     \begin{tabular}{l|l|r}
     \hline
      Model & Dataset & Correlation \\
      \hline\hline
      Classical Skip-gram  & WordSim353 & $72.5$\\
      Complex Skip-gram w/sigmoid  &   & $69.3$  \\
      Fitted PQC  &   & $69.3$  \\
      \hline
      Classical Skip-gram  & MEN  & $71.5$\\
      Complex Skip-gram w/sigmoid  &   & $71.3$  \\
      Fitted PQC   &    & $71.3$  \\
      \hline
      Classical Skip-gram  & RG-65  & $76.9$\\
      Complex Skip-gram w/sigmoid  &   & $76.1$  \\
      Fitted PQC   &    & $76.1$  \\
      \hline
      Classical Skip-gram  & SCWS  & $65.3$\\
      Complex Skip-gram w/sigmoid  &   & $65.9$  \\
      Fitted PQC   &    & $65.7$  \\
      \hline
\end{tabular}
\caption{Spearman correlation on various datasets for the large-scale models with an embedding dimension of 64 for complex, 100 for real-valued. 3 layers of A5 for the fitted PQC.}
\label{tab:more_data_2}
\end{center}
\end{table}

\section{Related Work}
\label{sec:related}

The idea of applying quantum-inspired models to lexical semantics goes back at least to the work of Widdows, who investigated using quantum logic for modelling negation \cite{widdows-2003-orthogonal}. 
Researchers have also investigated how the features of quantum mechanics, such as superposition and entanglement, can be applied to distributional semantics \cite{blacoe-etal-2013-quantum} and also for learning word embeddings \cite{li-etal-2018-quantum}. There is also a sub-field of Information Retrieval (IR) devoted to quantum-inspired models, based on the extensive use of vector-space models in IR \cite{van-rijsbergen-book,sordoni2013}. There is also a long history of studying complex-valued neural networks in machine learning \cite{142037,arjovsky_urnns,Trabelsi}. All of this work fits in the ``quantum-inspired'' regime, in that it was carried out in classical simulation with no indication of how the models could be run on quantum devices  (partly because such devices did not exist when the earlier work was written). 

NLP has also used quantum models more broadly \cite{wu-etal-2021-natural,widdows_survey}, again based on the observation that the 
mathematics of quantum vector spaces shares similarities with vector space models of semantics \cite{clark-lex-sem}. One particular focus has been on compositional distributional models of language \cite{coecke2010}, with a recent demonstration of solving a toy NLP problem on actual quantum hardware \cite{qnlp-practice}. Ref. \cite{sequence_processing_carys} investigates PQC-based architectures for sequence modelling, including some language tasks such as sentiment analysis, and also includes some test runs on quantum hardware.

In the Introduction we suggested that one potential advantage of using complex-valued embeddings is in learning hierarchical relationships. Here there has been some theoretical work on using a natural partial order on density matrices to model the hyponymy relation \cite{bankova}. Logical and conversational negation has also been studied in the context of quantum models \cite{lewis2020,rodatz-etal-2021-conversational}. Ref.~\cite{qoncepts} investigates how a category-theoretic treatment of conceptual spaces \cite{gardenfors2014} can be instantiated with a (hybrid) quantum model, showing how a CNN can be trained to predict the parameters of a PQC representing the colour, size, shape and position of a simple image.

\section{Conclusion and Future Work}
\label{sec:conclusion}

We have presented a variety of methods for training complex-valued word embeddings, including some physically-inspired variants which use PQCs. Overall the majority of the models perform competitively with a classical real-valued baseline on some standard similarity and relatedness datasets. In particular, the PQCs which have been trained to fit arbitrary complex vectors trained on a large corpus perform extremely well.

As mentioned, all of the experiments in this report have been run in simulation on classical hardware, but it is interesting to consider which parts could be run on quantum hardware (either now or potentially in the future). For training, only those methods which use PQCs for both the focal and the context words are amenable to quantum devices (since there is no efficient method for preparing an arbitrary complex embedding as a physical quantum state). However, there are a number of barriers to currently running such experiments in practice, especially at scale, and as the number of qubits used for the word representations increases. Aside from the current speed and reliability of the devices, one issue is that calculating gradients on quantum hardware is expensive, essentially because there is no quantum equivalent of the backpropagation algorithm \cite{qbackprop-neurips-2023}. Second, there is a large body of theory in QML which has identified and studied a phenomenon known as ``barren plateaus'' where, as the size of the quantum vector space increases exponentially with the number of qubits, the loss landscape becomes flat almost everywhere, making gradient-based learning extremely challenging \cite{barren_plateau}. Whether there are models which avoid barren plateaus, but still remain hard to simulate classically, is an ongoing research question \cite{barren_plateaus}.

A more immediate use of the word PQCs trained in simulation is in QNLP models at inference time. Classically-trained QNLP models have already been run on quantum hardware \cite{qnlp-practice,sequence_processing_carys} and an accurate quantum emulator \cite{xuQRNN}, in some cases with models which become hard to simulate classically as the size of the underlying vector space increases (which can happen at inference time through the use of models with particular compositional structure \cite{qdiscocirc}). The word PQCs we have trained could be run on quantum hardware and used as input to such models. Another area where there could be quantum advantage is in \emph{sampling} from a classically-trained generative model \cite{sequence_processing_carys}. Whether any of this increased expressivity results in improved performance in practice, however, remains to be seen.


There are many interesting directions for future work. The first is investigating alternative Skip-gram architectures, including the use of basis encoding described in Section~\ref{sec:pqc_encoding} which can be used to represent a distribution over the full vocabulary which can then be applied to a multi-class classification task as in the original Skip-gram model. Another direction is to consider alternative classical word embedding models, such as GloVe \cite{pennington-etal-2014-glove}. The GloVe model also uses an inner product between a focal and context word, but rather than being fed into a sigmoid for binary classification, the inner product is used to predict the log frequency counts of the word pair. Whether the overlap measure for complex vectors is suitable for such a task requires further theoretical and experimental investigation.

\section*{Acknowledgements}

Thanks to Wenduan Xu for preparing the large Wikipedia corpus; to Saskia Bruhn for help checking the gradient calculations; to Sara Sabrina Zemlji\v{c} for helpful discussions and for help with some of the hyperparameter tuning; and to Nikhil Khatri and Gabriel Matos for detailed and helpful feedback.

\bibliographystyle{alpha}
\bibliography{refs}

\appendix
\section{Ansatze from Sim et al.}
\label{sec:ansatze}

The ansatze used in the experiment in Section~\ref{sec:expts_pqc}, taken from Figure~2 in Ref.~\cite{expressive-pqc}:

\begin{figure}[h!]
    \centering
    \includegraphics[width=0.4\linewidth]{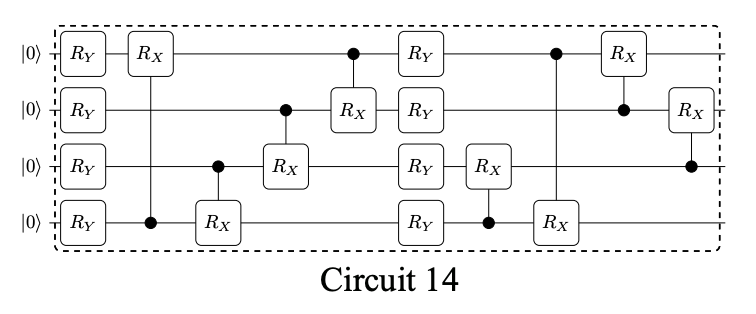}
    \includegraphics[width=0.5\linewidth]{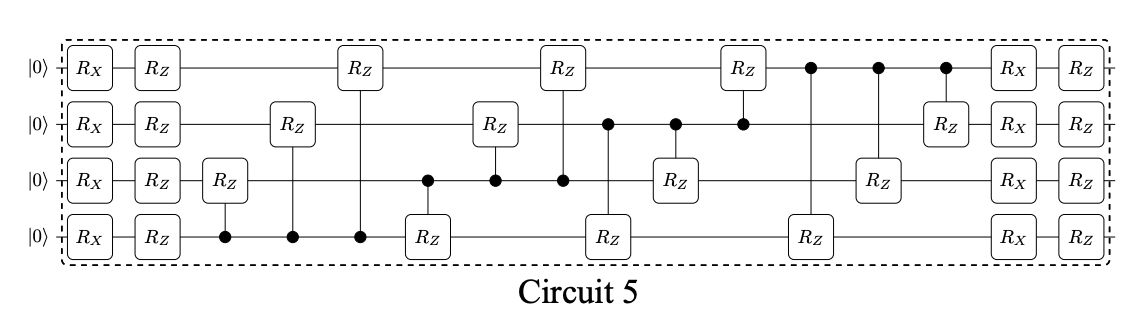}
    \includegraphics[width=0.4\linewidth]{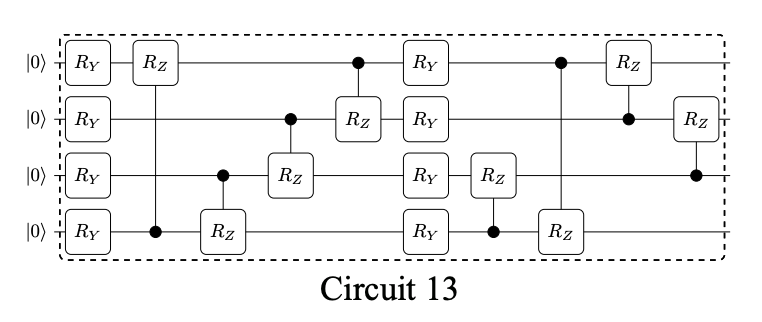}
    \includegraphics[width=0.3\linewidth]{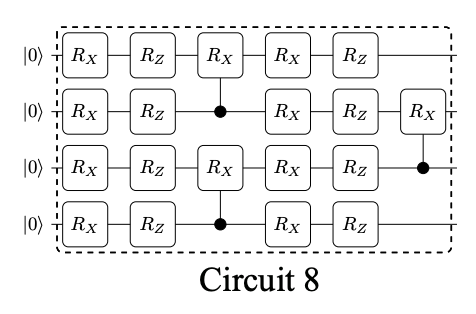}
    \includegraphics[width=0.3\linewidth]{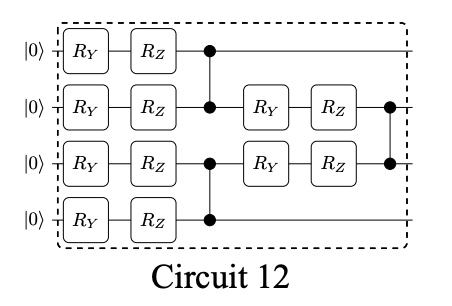}
    \includegraphics[width=0.3\linewidth]{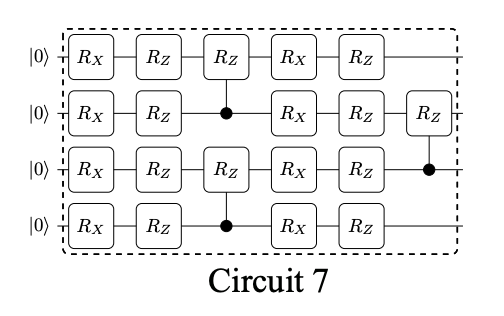}
    \includegraphics[width=0.4\linewidth]{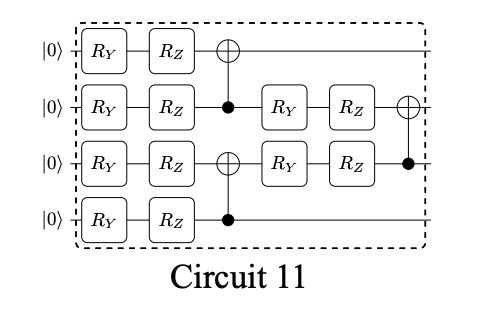}
    \includegraphics[width=0.3\linewidth]{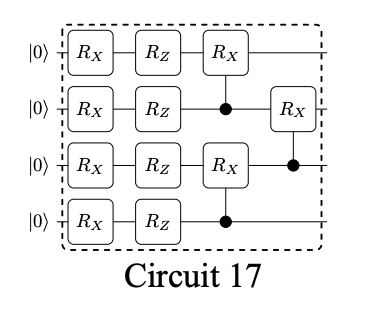}
    \includegraphics[width=0.25\linewidth]{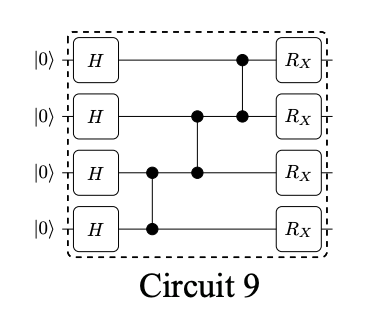}
\end{figure}

\end{document}